\documentclass{article}
\usepackage{spconf,amsmath,graphicx,hyperref}

\usepackage{booktabs}   % For professional table rules
\usepackage{multirow, makecell}   % For multi-row cells
\usepackage{graphicx}   % Often needed for other things, good to have
\usepackage{mathtools,amssymb}

\title{AutoVQA-G: Self-Improving Agentic Framework for Automated Visual Question Answering and Grounding Annotation}

\name{Rongsheng Hu$^{1}$ \qquad Runwei Guan$^{2}$ \qquad Yicheng Di$^{1}$ \qquad Jiayu Bao$^{1}$ \qquad Yuan  Liu$^{1}$\sthanks{Corresponding author is Yuan Liu. Accepted at IEEE ICASSP 2026.}}

\address{$^{1}$ School of Artificial Intelligence and Computer Science, Jiangnan University, Wuxi, China \\
    $^{2}$ Thrust of Artificial Intelligence, HKUST(GZ), Guangzhou, China \\
    \textit{hu\_rongsheng@stu.jiangnan.edu.cn, runwayrwguan@hkust-gz.edu.cn,} \\
    \textit{\{diyicheng1, 7243115008\}@stu.jiangnan.edu.cn, lyuan1800@jiangnan.edu.cn}}

\begin{document}
\ninept
\maketitle
%
%High-quality Visual Question Answering with Grounding (VQA-G) datasets, which require models to answer a visual question and localize the supporting evidence, are crucial for benchmarking Vision-Language Models (VLMs), yet manual annotation is unscalable. Current automated methods attempt to fill this gap but suffer from two critical flaws: \textbf{(1)} inconsistent data fidelity due to model hallucinations; \textbf{(2)} a lack of robust verification, often relying on brittle, heuristic checks. To address these challenges, we introduce AutoVQA-G, a self-improving agentic framework for automated VQA-G annotation. AutoVQA-G tackles these issues head-on through an iterative refinement loop. To combat unreliable verification, it incorporates a Consistency Evaluation module, which uses Chain-of-Thought (CoT) visual reasoning to perform a deep, step-by-step assessment of generated data. To systematically improve data fidelity at its source, a memory-augmented Prompt Optimization agent analyzes the critiques from failed cases to dynamically refine and enhance the generation rubrics for subsequent iterations. Our experiments demonstrate that AutoVQA-G is a scalable and cost-effective solution that yields accurately grounded VQA-G datasets with higher quality than state-of-the-art Multimodal LLMs, thereby facilitating more robust training and evaluation of next-generation VLMs. We release codes at \url{https://github.com/rohnson1999/AutoVQA-G}.
\begin{abstract}
Manual annotation of high-quality visual question answering with grounding (VQA-G) datasets, which pair visual questions with evidential grounding, is crucial for advancing vision-language models (VLMs), but remains unscalable. Existing automated methods are often hindered by two key issues: \textbf{(1)} inconsistent data fidelity due to model hallucinations; \textbf{(2)} brittle verification mechanisms based on simple heuristics. To address these limitations, we introduce AutoVQA-G, a self-improving agentic framework for automated VQA-G annotation. AutoVQA-G employs an iterative refinement loop where a Consistency Evaluation module uses Chain-of-Thought (CoT) reasoning for fine-grained visual verification. Based on this feedback, a memory-augmented Prompt Optimization agent analyzes critiques from failed samples to progressively refine generation prompts. Our experiments show that AutoVQA-G generates VQA-G datasets with superior visual grounding accuracy compared to leading multimodal LLMs, offering a promising approach for creating high-fidelity data to facilitate more robust VLM training and evaluation. We release codes publicly\footnote{\url{https://github.com/rohnson1999/AutoVQA-G}}.
\end{abstract}

\section{Introduction}
\label{sec:intro}
The advancement of sophisticated vision-language models (VLMs) is fundamentally tied to the availability of large-scale, high-quality datasets that provide fine-grained supervision~\cite{alayrac2022flamingo, hurst2024gpt, comanici2025gemini}. Among these, datasets combining visual question answering (VQA) and visual grounding (VG) are particularly valuable, as they foster deeper visual reasoning and enhance model interpretability by linking semantic concepts to specific image regions~\cite{hudson2019gqa, kamath2021mdetr}. However, the manual creation of such VQA with grounding (VQA-G) datasets is expensive, time-consuming, and difficult to scale, especially in specialized domains requiring expert knowledge~\cite{zhu2016visual7w, Krishna2017}.

To mitigate these challenges, automated data annotation using VLMs has emerged as a promising alternative. This trend spans from general-purpose labeling platforms~\cite{ ghazouali2025visiofirm} to specialized applications in autonomous driving~\cite{zhou2024openannotate2} and medical imaging~\cite{fink2023potential}. Within the VQA-G domain, early efforts such as VQ²A demonstrated the feasibility of generating VQA pairs from image captions~\cite{changpinyo-etal-2022-may}, while other pipelines focused on creating large-scale instruction-following data~\cite{liu2023visual, chung2024scaling}. More recently, the focus has shifted towards enhancing data quality and cognitive complexity. Researchers have developed automated pipelines to synthesize high-quality grounding data with Chain-of-Thought (CoT) reasoning~\cite{wei2022chain} to improve VLMs perception~\cite{ma2025deepperception, bai2025univg}. In parallel, recent advances have seen the rise of agentic frameworks that tackle annotation through complex cognitive reasoning~\cite{faure2025moviecore} and multi-agent cooperation in specialized domains~\cite{ALGPT10681241}, pushing the boundaries of automation.

Despite this progress, current automated methods are limited by two key challenges. First, as single-pass systems, they are susceptible to VLMs' hallucinations~\cite{li2023evaluating}, yielding annotations of inconsistent fidelity that demand costly manual correction. The second challenge is brittle verification. Instead of a deep assessment of the visual content, many systems, including agentic frameworks, rely on fragile, heuristic checks for internal validation~\cite{zhou2023lima}. This approach frequently fails in complex or out-of-distribution scenarios where such predefined rules are no longer applicable.

To overcome these limitations, we introduce \textbf{AutoVQA-G}, a self-improving agentic framework built on a dynamic generate, evaluate, and refine loop. It tackles brittle verification via a Consistency Evaluation module that uses CoT reasoning~\cite{wei2022chain} to produce detailed, step-by-step visual critiques and verifiable quality scores. To combat inconsistent data quality, a memory-augmented Prompt Optimization agent analyzes evaluation feedback, identifies failure patterns, and iteratively refines generation rubrics, allowing the system to learn from errors and improve over time. The loop runs until annotations achieve high consistency, at which point they’re accepted as high-fidelity data.

\begin{figure*}[t!]
  \centering
  \includegraphics[width=\linewidth]{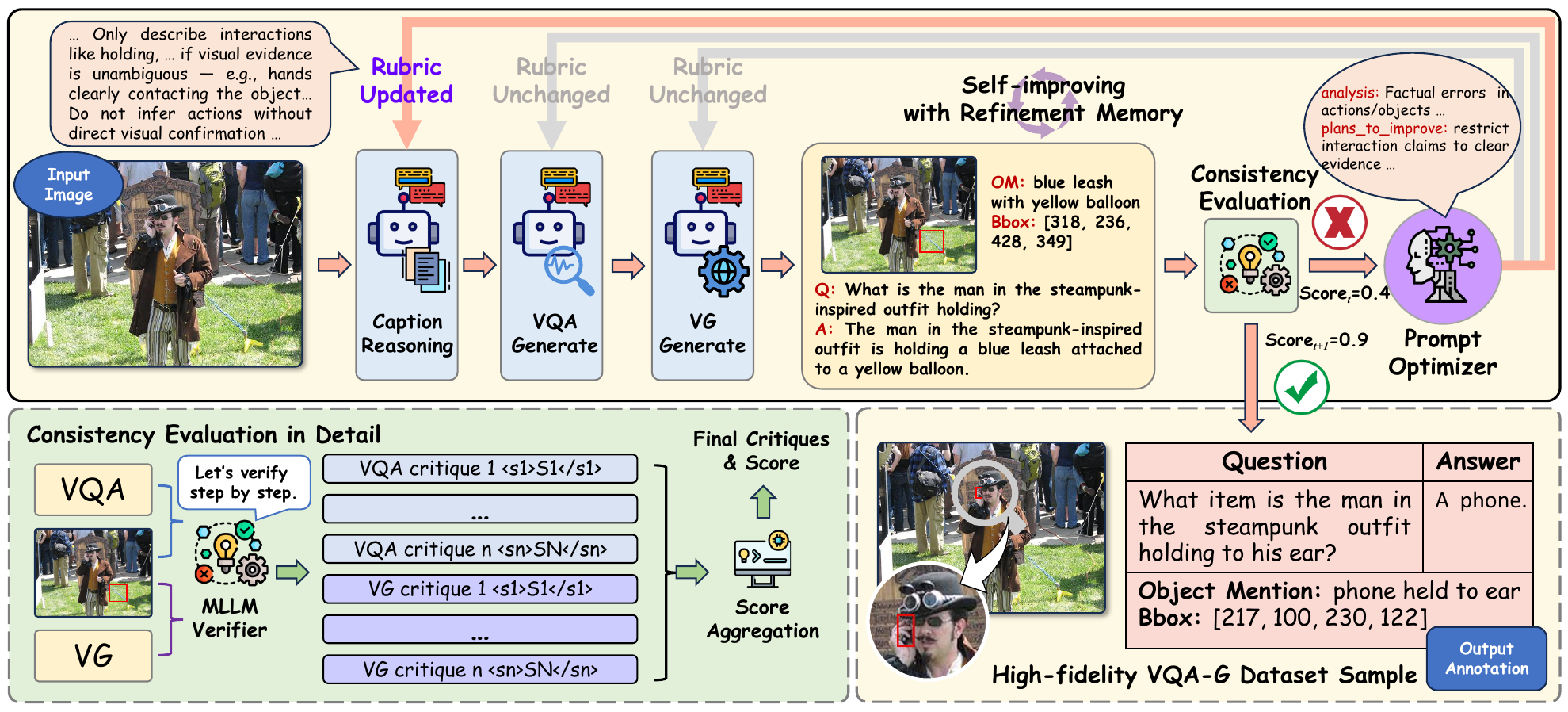}
  \vspace{-0.25in} % Reduce space between image and caption
  \caption{Overview of the AutoVQA-G automated annotation framework, which iteratively refines VQA-G data for an input image. An initial, low-quality draft (e.g., about a ``balloon leash") is rejected by the \textbf{Consistency Evaluation} module (Score=0.4). The \textbf{Prompt Optimization} agent uses the critique to update the generation prompts, allowing the framework to produce a corrected, high-fidelity sample (about a ``phone") in the next attempt, which passes verification and is accepted as the output annotation.}
  \label{fig:vqa-grounding-models}
  %\vspace{-0.1in} % Reduce space after the figure*
\end{figure*}

The complete workflow of AutoVQA-G is illustrated in Fig.~\ref{fig:vqa-grounding-models}. Its modular architecture not only enables the creation of fully aligned VQA-G data but also supports the independent generation of high-quality VQA or VG annotations, making AutoVQA-G a versatile toolkit. Our main contributions are summarized as follows:
\begin{enumerate}
    \item We propose AutoVQA-G, a novel agentic framework that automates VQA-G annotation through a self-improving, iterative refinement loop.
    \item We introduce a CoT-based Consistency Evaluation module for fine-grained, interpretable VQA-G verification, and a Prompt Optimization agent with memory of past attempts and dynamic routing for targeted rubric updates.
    \item We demonstrate through extensive experiments that AutoVQA-G outperforms leading VLMs in quality and consistency, offering a scalable, cost-effective solution.
\end{enumerate}

%================================================================
% Section 2: The AutoVQA-G Framework 
%================================================================

\section{The AutoVQA-G Framework}
\label{sec:method}

We introduce \textbf{AutoVQA-G}, a self-improving agentic framework that iteratively generates high-fidelity VQA-G datasets via generate–evaluate–refine cycles (Fig.~\ref{fig:vqa-grounding-models}, \S\S~\ref{subsec:generation}–\ref{subsec:refinement}). 

%----------------------------------------------------------------
\subsection{Modular VQA-G Annotation Generation}
\label{subsec:generation}

In the generation stage at each iteration $t$, a candidate annotation draft, denoted as $D_t$, is constructed through a structured, multi-step pipeline. Given an input image $I \in \mathbb{R}^{H \times W \times 3}$ and generation rubrics $\mathbf{R}^{(t)} = \{R_{\text{cap}}^{(t)}, R_{\text{vqa}}^{(t)}, R_{\text{vg}}^{(t)}\}$, each generation step is modeled as a conditional probability distribution parameterized by distinct VLM components.

The Caption Reasoning module generates structured semantic context:
\begin{equation}
C_R \sim p(c \mid I, R_{\text{cap}}^{(t)}; \theta_{\text{cap}}),
\label{eq:caption_reasoning}
\end{equation}
where $\theta_{\text{cap}}$ denotes the VLM parameters for captioning, and $C_R$ encodes object attributes and spatial relationships.

The VQA Generation module incorporates visual and textual features to generate question-answer pairs:
\begin{equation}
(q_t, a_t) \sim p(q, a \mid I, C_R, R_{\text{vqa}}^{(t)}; \theta_{\text{vqa}}),
\label{eq:vqa_generation}
\end{equation}
where $\theta_{\text{vqa}}$ are the parameters for the VQA module.

The VG Generation module performs two-stage localization. In the first stage, it generates an object mention:
\begin{equation}
m_t \sim p(m \mid I, q_t, a_t, R_{\text{vg}}^{(t)}; \theta_{\text{vg}}),
\label{eq:mention_generation}
\end{equation}
where $\theta_{\text{vg}}$ parametrizes mention grounding. The second stage uses a deterministic function to map the mention to spatial coordinates:
\begin{equation}
b_t = \arg\max_{b \in \mathcal{B}} p(b \mid I, m_t; \theta_{\text{ground}}),
\label{eq:grounding}
\end{equation}
where $\mathcal{B} \subseteq \mathbb{R}^4$ is the space of valid bounding boxes, and $\theta_{\text{ground}}$ are localization parameters. 

The complete draft is $D_t = \{(q_t, a_t), (m_t, b_t)\}$.

%----------------------------------------------------------------
\subsection{Chain-of-Thought Consistency Verification}
\label{subsec:evaluation}

Recent work shows that LLM-as-judge with CoT~\cite{wei2022chain} reasoning enables effective inference-time feedback for training reasoning-capable models~\cite{liu2025inference}. We introduce CoT-based verifiers to assess draft quality in AutoVQA-G.

Each draft $D_t$ is assessed using specialized CoT vision verifiers, which decompose verification into interpretable reasoning steps. We employ two verifiers: $E_{\text{vqa}}$ for VQA consistency and $E_{\text{vg}}$ for visual grounding consistency.

For VQA consistency, the verifier outputs step-wise assessments:
\begin{equation}
\{(c_{i}^{\text{vqa}}, s_{i}^{\text{vqa}})\}_{i=1}^{n_{\text{vqa}}} = E_{\text{vqa}}(I, q_t, a_t),
\end{equation}
where $c_{i}^{\text{vqa}}$ is the $i$-th reasoning step with a corresponding score $s_{i}^{\text{vqa}} \in [0,1]$, and $n_{\text{vqa}}$ is the total reasoning steps for VQA.

For visual grounding, the verifier assesses the image $\hat{I}_t = I \oplus b_t$, where $\oplus$ denotes superimposing the bounding box $b_t$ over $I$:
\begin{equation}
\{(c_{i}^{\text{vg}}, s_{i}^{\text{vg}})\}_{i=1}^{n_{\text{vg}}} = E_{\text{vg}}(\hat{I}_t, m_t),
\end{equation}
where $c_{i}^{\text{vg}}$ and $s_{i}^{\text{vg}} \in [0,1]$ are the grounding reasoning step and its score, and $n_{\text{vg}}$ is the number of grounding reasoning steps.

The aggregate consistency score, $S_t$, combines both modalities:
\begin{equation}
S_t = w_{\text{vqa}} \cdot \underbrace{\frac{1}{n_{\text{vqa}}} \sum_{i=1}^{n_{\text{vqa}}} s_{i}^{\text{vqa}}}_{S_{t}^{\text{vqa}}} + w_{\text{vg}} \cdot \underbrace{\frac{1}{n_{\text{vg}}} \sum_{i=1}^{n_{\text{vg}}} s_{i}^{\text{vg}}}_{S_{t}^{\text{vg}}},
\end{equation}
where \( w_{\text{vqa}} \) and \( w_{\text{vg}} \) are weighting coefficients in \([0,1]\) that sum to one. The complete critique is given by  
\begin{equation}
C_t = \text{concat}\big(\{c_{i}^{\text{vqa}}\}, \{c_{i}^{\text{vg}}\}\big).
\end{equation}

Acceptance is based on a threshold $\tau$:
\begin{equation}
D_{\text{accept}} = 
\begin{cases}
D_{\text{accept}} \cup \{D_t\} & \text{if } S_t \geq \tau \\
D_{\text{accept}} & \text{otherwise}
\end{cases}.
\end{equation}

%----------------------------------------------------------------
\subsection{Memory-Augmented Prompt Optimization}
\label{subsec:refinement}

Inspired by recent LLM-based optimization frameworks that iteratively refine prompts using feedback signals~\cite{yang2023large,manas2024improving}, we design a memory-augmented agent to systematically improve generation rubrics while avoiding redundant or cyclic updates. 

If $S_t < \tau$, a Prompt Optimization agent refines the generation rubrics using observed failure signals. The agent maintains a historical memory $\mathcal{H}^{(t)} = \{(D_i, C_i, \mathbf{R}^{(i)})\}_{i=0}^{t}$, recording each draft, critique, and rubric set, to prevent cyclic updates and promote consistent improvement.

Reasoning and action generation proceed as:
\begin{equation}
z_t \sim p(z \mid \mathbf{R}^{(t)}, D_t, C_t, \mathcal{H}^{(t-1)}; \phi),
\end{equation}
where $\phi$ denotes the parameters of a pre-trained large language model, and $z_t$ encodes both failure diagnosis and a corrective plan.

A parsing function $\pi: z_t \mapsto (k^*, \Delta R_{k^*}^{(t)})$, where $k^* \in \{\text{cap}, \text{vqa}, \text{vg}\}$, extracts the target module and the corresponding refinement instruction $\Delta R_{k^*}^{(t)}$. The rubrics are updated as:
\begin{equation}
R_{k}^{(t+1)} = 
\begin{cases}
R_{k}^{(t)} \uplus \Delta R_{k}^{(t)} & \text{if } k = k^* \\
R_{k}^{(t)} & \text{otherwise}
\end{cases},
\end{equation}
for $k \in \{\text{cap}, \text{vqa}, \text{vg}\}$, where $\uplus$ denotes the rubric update operation that incorporates the refinement.

This targeted, memory-driven refinement mechanism enables AutoVQA-G to iteratively enhance annotation quality, halting either when $S_t \geq \tau$ or upon reaching a maximum number of iterations.

%================================================================
% Section 3: Experiments
%================================================================
\section{Experiments}
\label{sec:experiments}

%----------------------------------------------------------------
\subsection{Experimental Settings}
\label{subsec:protocol}

\subsubsection{Implementation and Datasets}
AutoVQA-G is a training-free framework implemented with a suite of publicly available models. For our experiments, the generation (MiniCPM-o 2.6\footnote{https://huggingface.co/openbmb/MiniCPM-o-2\_6}), localization (GroundingDINO\footnote{https://huggingface.co/IDEA-Research/grounding-dino-base}) models and all evaluations are run locally on four NVIDIA RTX 4090 GPUs. The CoT verifier (Qwen2.5-VL 72B\footnote{https://huggingface.co/Qwen/Qwen2.5-VL-72B-Instruct}) and prompt optimizer (DeepSeek V3\footnote{https://huggingface.co/deepseek-ai/DeepSeek-V3-0324}) are accessed via API. The self-improving loop terminates when the consistency score exceeds $\tau=0.9$ (a weighted sum with $w_{\text{vqa}}=0.7$, $w_{\text{vg}}=0.3$) or after five iterations, with the best result being chosen. These hyperparameters were tuned on a validation set. Experiments are conducted on 10,000 images from the tellingQA subset of Visual7W~\cite{zhu2016visual7w} and VizWiz-VQA-G~\cite{gurari2018vizwizgrandchallengeanswering} (hereafter “VizWiz”), where VizWiz segmentation masks are converted to bounding boxes for consistency. All prompt templates are publicly available\footnote{https://github.com/rohnson1999/AutoVQA-G}.

\begin{table}[hbt!]
\centering
\caption{Statistics of samples generated by AutoVQA-G.}
\label{tab:framework_stats}
\resizebox{\columnwidth}{!}{%
\begin{tabular}{lcc}
\toprule
\textbf{Metric} & \textbf{Visual7W} & \textbf{VizWiz} \\
\midrule
\multicolumn{3}{l}{\textit{\textbf{Efficiency and Cost}}} \\
Success Rate (\%) & 91.8 & 89.2 \\
Avg. Iterations per Success & 1.62 & 2.15 \\
Avg. Total Tokens per Success (K) & 2.1 & 3.1 \\
\midrule
\multicolumn{3}{l}{\textit{\textbf{Generated Content Characteristics}}} \\
Avg. Question Length (words) & 17.39 & 10.96 \\
Avg. Answer Length (words) & 12.13 & 8.83 \\
Avg. Object Mention Length (words) & 12.61 & 9.69 \\
Avg. BBox Area (\% of Image) & 19.7 & 45.2 \\
\midrule
\multicolumn{3}{l}{\textit{\textbf{Question Complexity Distribution (\%)}}} \\
\quad - Relational \& Counting & 47.4 & 38.5 \\
\quad - Attribute \& Other & 52.6 & 61.5 \\
\bottomrule
\end{tabular}%
}
\end{table}

\begin{figure}[hbt!]
   \centering
   \includegraphics[width=\linewidth]{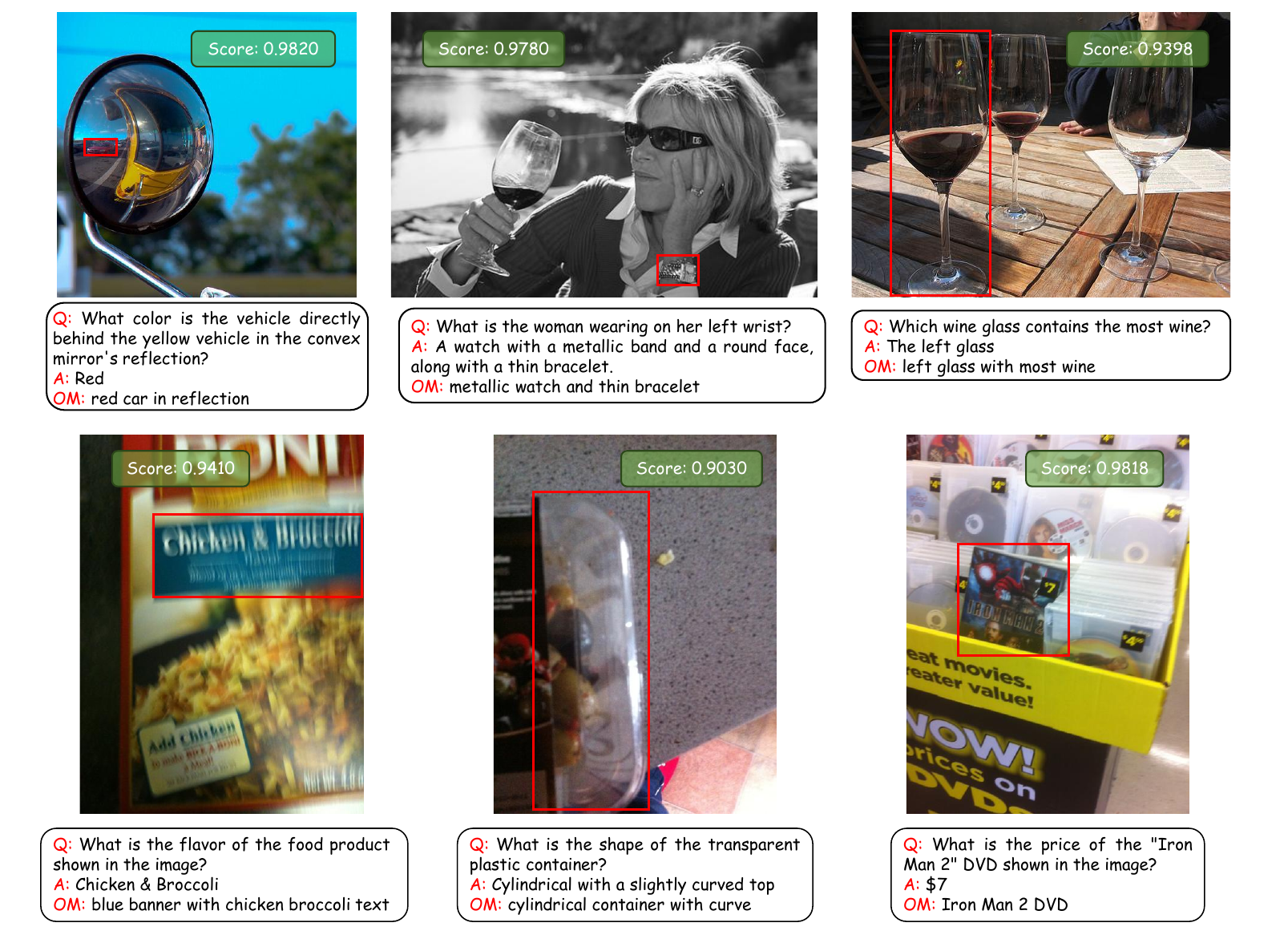}
   \vspace{-0.25in}
   \caption{Qualitative examples generated by AutoVQA-G. The framework successfully produces high-consistency data across diverse scenarios, showcasing complex reasoning in QA pairs and precise, fine-grained visual grounding. (Better viewed zoomed in.)}
    \label{fig:qualitative_examples}
    %\vspace{-0.1in}
\end{figure}

\subsubsection{Evaluation Protocol}
We benchmark AutoVQA-G against GPT-4o~\cite{hurst2024gpt} and Gemini 2.5 Flash~\cite{comanici2025gemini}. For a fair comparison of annotation strategies, we isolate the impact of our iterative loop versus single-pass generation by equipping both AutoVQA-G and a GPT-4o (ZS, tool-assisted) baseline with the same external grounding module (GroundingDINO). Other baselines rely on their native grounding capabilities. VQA quality is measured via VQAScore~\cite{lin2024evaluating}, TIFA~\cite{hu2023tifa}, and CLIPScore~\cite{hessel2021clipscore}. For an objective visual grounding evaluation, 10 annotators re-annotated 6,000 samples (500 per method/dataset), enabling a fair comparison via mIoU and Acc@0.5IoU. The final VQA-G score averages the mean VQA and visual grounding metrics.

\begin{table*}[t!]
\centering
\caption{Performance comparison against state-of-the-art VLM baselines on Visual7W and VizWiz. AutoVQA-G demonstrates a superior balance of VQA quality and visual grounding accuracy, substantially outperforming all baselines in grounding-specific metrics (mIoU, Acc@0.5IoU) and achieving the highest overall VQA-G score. Best results are in \textbf{bold}.}
\label{tab:main_results}
\resizebox{\textwidth}{!}{%
\begin{tabular}{llccccccc}
\toprule
\multirow{2}{*}{\textbf{Dataset}} & 
\multirow{2}{*}{\textbf{Method}} & 
\multicolumn{3}{c}{\textbf{VQA Evaluation}} & 
\multicolumn{2}{c}{\textbf{VG Evaluation}} & 
\multirow{2}{*}{\textbf{\makecell{Average \\ VQA-G Score}} ($\uparrow$)} \\
\cmidrule(lr){3-5} \cmidrule(lr){6-7}
& & CLIPScore ($\uparrow$) & TIFA ($\uparrow$) & VQAScore ($\uparrow$) & mIoU ($\uparrow$) & Acc@0.5IoU ($\uparrow$) & \\
\midrule
\multirow{6}{*}{Visual7W} 
& Human Annotation & 0.651 & 0.865 & 0.890 & 0.517 & 0.560 & 0.670 \\
& GPT-4o (ZS, tool-assisted) & 0.733 & \textbf{0.908} & \textbf{0.923} & 0.455 & 0.510 & 0.669 \\
& GPT-4o (CoT) & \textbf{0.738} & 0.903 & 0.918 & 0.334 & 0.240 & 0.570 \\
& Gemini (ZS) & 0.708 & 0.877 & 0.894 & 0.390 & 0.440 & 0.621 \\
& Gemini (CoT) & 0.719 & 0.894 & 0.899 & 0.382 & 0.400 & 0.614 \\
& \textbf{AutoVQA-G} & 0.735 & 0.819 & 0.896 & \textbf{0.634} & \textbf{0.720} & \textbf{0.747} \\
\midrule
\multirow{6}{*}{VizWiz} 
& Human Annotation & 0.724 & 0.794 & 0.762 & 0.627 & 0.640 & 0.697 \\
& GPT-4o (ZS, tool-assisted) & 0.753 & \textbf{0.849} & \textbf{0.907} & 0.472 & 0.525 & 0.667 \\
& GPT-4o (CoT) & 0.754 & 0.841 & 0.901 & 0.354 & 0.340 & 0.590 \\
& Gemini (ZS) & 0.745 & 0.818 & 0.883 & 0.445 & 0.480 & 0.639 \\
& Gemini (CoT) & 0.745 & 0.819 & 0.884 & 0.333 & 0.340 & 0.576 \\
& \textbf{AutoVQA-G} & \textbf{0.757} & 0.800 & 0.874 & \textbf{0.649} & \textbf{0.680} & \textbf{0.737} \\
\bottomrule
\end{tabular}%
}
\end{table*}

%----------------------------------------------------------------
\begin{table}[hbt!]
\centering
\caption{Ablation study of the AutoVQA-G framework on Visual7W. We evaluate the impact of removing key components on VQA quality (VQAScore) and grounding accuracy (mIoU).}
\label{tab:ablation}
\resizebox{\columnwidth}{!}{%
\begin{tabular}{lcc}
\toprule
\textbf{Configuration} & \textbf{VQAScore} ($\uparrow$) & \textbf{mIoU} ($\uparrow$) \\
\midrule
\textbf{AutoVQA-G (Full Model)} & \textbf{0.896} & \textbf{0.634} \\
\midrule
(1) Single-Pass Generation (w/o loop) & 0.863 & 0.380 \\
(2) w/ Score-only Verification (no CoT) & 0.879 & 0.495 \\
(3) w/o Dynamic Routing & 0.875 & 0.561 \\
(4) w/o Memory & 0.885 & 0.582 \\
\bottomrule
\end{tabular}%
}
\end{table}
%----------------------------------------------------------------
\subsection{Results and Analysis}
\label{subsec:results}

\subsubsection{Operational Analysis}
As detailed in Table~\ref{tab:framework_stats}, AutoVQA-G achieves strong success rates on both Visual7W (91.8\%) and the more challenging VizWiz (89.2\%). The framework adapts its effort to input difficulty, requiring more iterations (1.62 vs. 2.15) and thus more tokens for VizWiz. The generated content is also tailored to the dataset's nature; annotations for the richer scenes in Visual7W feature more complex questions and more precise grounding on smaller object details (19.7\% avg. bbox area vs. 45.2\% for VizWiz). Critically, for both datasets, AutoVQA-G generates a high percentage of cognitively demanding relational and counting questions, proving its ability to produce high-value training data beyond simple attribute queries.

\subsubsection{Main Quantitative Results}
As shown in Table~\ref{tab:main_results}, AutoVQA-G achieves the highest overall VQA-G score, demonstrating the value of its iterative process. Critically, while equipping GPT-4o with an external tool (GPT-4o ZS, tool-assisted) substantially improves its grounding scores and establishes a strong baseline, AutoVQA-G still outperforms it by a significant margin. This result demonstrates that our performance gain stems not from the tool itself, but from the agentic process of evaluating, critiquing, and refining annotations. Our loop effectively filters inconsistencies that a single-pass approach, even when tool-augmented, cannot address.

A notable finding is that AutoVQA-G's visual grounding scores exceeded those obtained from the re-evaluation of the original human annotations, functioning as an effective consistency enforcer. The iterative refinement loop systematically selects for less ambiguous queries and applies a uniform quality standard via its CoT verifier. This process results in a highly consistent dataset. Impressively, our framework pairs this superior grounding fidelity with competitive VQA quality, rivaling that of leading VLMs like GPT-4o. This is particularly noteworthy given that AutoVQA-G relies on a much smaller 8B-scale VLM for initial draft generation. This demonstrates that our agentic framework effectively elevates the capabilities of smaller models, offering a resource-efficient and powerful approach for automated annotation.

\subsubsection{Ablation and Qualitative Analysis}
To dissect the contribution of each component, we conduct an ablation study on the Visual7W dataset (Table~\ref{tab:ablation}). The results indicate that each element of the framework contributes to the final performance, omitting any one results in a performance drop. The most substantial impact occurs in the single-pass configuration (1), which effectively removes the agent-driven refinement loop, causing the mIoU to fall to 0.380. This highlights the critical role of the iterative process. Furthermore, the quality of the feedback signal proved important; removing the verifier's CoT reasoning (2) also leads to a significant drop in mIoU. Lesser, yet notable, performance decreases occur when removing agent-specific mechanisms like dynamic routing (3) or memory (4), confirming their utility in enabling an efficient optimization strategy.

Qualitatively, Fig.~\ref{fig:qualitative_examples} showcases AutoVQA-G's ability to generate diverse and highly consistent data. The examples demonstrate the framework's capacity for complex reasoning (e.g., nuanced spatial analysis in a mirror's reflection) and robust, fine-grained localization, precisely pinpointing small targets like a watch or reading blurry text via OCR. The high scores across these challenging scenarios validate the effectiveness of our proposed method.

%----------------------------------------------------------------
\section{Conclusion}
\label{sec:conclusion}
We introduce AutoVQA-G, a self-improving agentic framework that replaces the error-prone single-pass annotation with an iterative ``generate-evaluate-refine'' loop, driven by verification of visual CoT consistency and memory-augmented prompt optimization. Experiments show it produces more consistent, accurately grounded VQA-G data, setting a new standard for automated annotation. While current computational overhead presents a limitation, AutoVQA-G offers a robust blueprint for using agentic systems to solve the critical data bottleneck in vision-language research.

\vfill
\pagebreak

\footnotesize

\section{ACKNOWLEDGMENTS}
\label{sec:acknow}

This work was supported by the National Natural Science Foundation of China (Grant No. 62472200).
% -------------------------------------------------------------------------
%\small
%\bibliographystyle{IEEEbib}
%\bibliography{strings,refs}
\bibliographystyle{IEEEtran}
\bibliography{strings,refs}

\end{document}